\documentclass[letterpaper, 10 pt, conference]{ieeeconf}  

\IEEEoverridecommandlockouts                              
\usepackage{graphicx}      
\usepackage{algorithm} 
\usepackage{algpseudocode} 
\usepackage{varwidth} 
\usepackage{cite}
\usepackage{diagbox}
\usepackage{subcaption}
\usepackage{amsfonts}
\usepackage{amsmath}
\usepackage{url}
\usepackage{gensymb}
\usepackage{tabularx,caption}
\usepackage[dvipsnames]{xcolor}
\usepackage{hyperref}

\newcommand{\argmax}{\mathop{\mathrm{argmax}}}  
\graphicspath{{figs/}}
\title{\LARGE \bf
Multi-level Reasoning for Robotic Assembly: \\ 
From Sequence Inference to Contact Selection

}

\author{Xinghao Zhu$^{1*}$, Devesh K. Jha$^{2}$, Diego Romeres$^{2}$, Lingfeng Sun$^{1}$, Masayoshi Tomizuka$^{1}$, Anoop Cherian$^{2\dagger}$
\thanks{$^{1}$Mechanical Systems Control Lab, UC Berkeley, Berkeley, CA, USA
{\tt\small \{zhuxh, lingfengsun, tomizuka\}@berkeley.edu}
}
\thanks{$^{2}$Mitsubishi Electric Research Laboratories (MERL), Cambridge, MA, USA
{\tt\small \{jha, romeres, cherian\}@merl.com}}
\thanks{$^*$Work done during MERL internship. $^\dagger$Corresponding author.}{}
}

\begin{document}
\maketitle
\thispagestyle{empty}
\pagestyle{empty}

\begin{abstract}
Automating the assembly of objects from their parts is a complex problem with innumerable applications in manufacturing, maintenance, and recycling. Unlike existing research, which is limited to target segmentation, pose regression, or using fixed target blueprints, our work presents a holistic multi-level framework for part assembly planning consisting of part assembly sequence inference, part motion planning, and robot contact optimization. We present the Part Assembly Sequence Transformer (PAST) -- a sequence-to-sequence neural network -- to infer assembly sequences recursively from a target blueprint. We then use a motion planner and optimization to generate part movements and contacts. To train PAST, we introduce D4PAS: a large-scale \emph{Dataset for Part Assembly Sequences} consisting of physically valid sequences for industrial objects.  Experimental results show that our approach generalizes better than prior methods while needing significantly less computational time for inference.  Further details on our experiments and results are available in the \href{https://www.youtube.com/watch?v=XNYkWSHkAaU&ab_channel=MitsubishiElectricResearchLabs%28MERL%29}{supplementary video}.

\end{abstract} 


\section{Introduction}
\label{sec: introduction}

The assembly of parts in accordance with a target blueprint presents a compelling research frontier in the domains of robotics and machine learning. This task not only represents a highly valuable functionality that autonomous robots can perform, but it also embodies a complex problem space characterized by indeterminate intricacies. 
Achieving successful assembly requires robots to master several complex skills: understanding part geometries, reasoning about physical interactions and collisions, and executing assembly plans with robust sensing capabilities.
To navigate through these complexities, robots need to cultivate an array of diverse competencies conducive to a successful assembly process. These include deciphering the assembly sequence, coordinating part trajectories, identifying points of contact, and their physical execution. Equally important is the robot's ability to generalize these competencies across various assemblies.

Previous investigations in robotics and computer vision have tackled this multifaceted challenge of part assembly through diverse methodological lenses.
For example, one line of research attempts to sidestep the complexities of physics, electing to focus on more specialized tasks, such as the segmentation of target blueprints~\cite{li2023general, li2020impartass} or the estimation of part poses~\cite{HuangZhan2020PartAssembly, NSM}.
A second vein of research emphasizes the importance of physical interactions, specializing in the assembly of predetermined targets~\cite{narang2022factory, 9482981, 9811973, 9561646}.
A third category adopts different target blueprints but imposes simplifying assumptions on part geometries (i.e., blocks)~\cite{Ghasemipour2022BlocksAL} or restricts its scope to a seen set of blueprint categories (e.g., chairs)~\cite{yu2021roboassembly}.
The use of reinforcement learning (RL) has seen success in this space~\cite{Ghasemipour2022BlocksAL, yu2021roboassembly}. However, RL-based solutions face challenges in terms of computational resources and efficiency. For instance, Ghasemipour et al.~\cite{Ghasemipour2022BlocksAL} requires an elaborate computational infrastructure involving thousands of CPUs and billions of steps for training, raising concerns about the practicality of the system.
Our work aims to advance the field of robotic assembly by holistically considering intricate physical interactions between parts and designing a supervised training paradigm, while our approach is applicable to a broad spectrum of practical target blueprints.

\begin{figure}[tb]
\begin{center}
\includegraphics[width=3.4in]{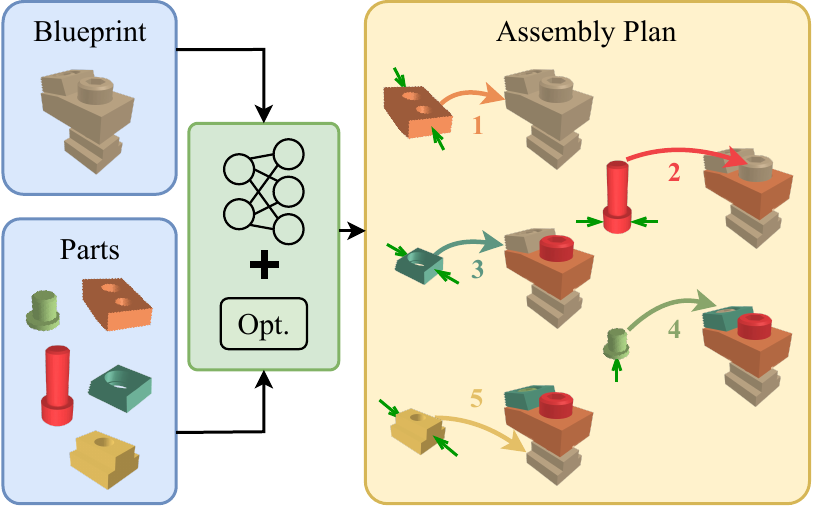}
\caption{
Our goal is to facilitate robotic assembly across different target blueprints. Utilizing point clouds from target blueprints and assembly parts, our method identifies feasible assembly sequences (indicated by colored numbers), orchestrates part motions (represented by colored long arrows), and pinpoints contact points (denoted by short green arrows).
}
\label{fig: teaser}
\vspace{-0.5em}
\end{center}
\end{figure}

To achieve successful robotic assembly, this study breaks down the task into three distinct and key sub-tasks: 1) inferring the sequence in which the parts should be assembled, guided by the target blueprint, part shapes, and assembled poses; 2) coordinating the movements of the individual parts; and 3) identifying viable contact points for robotic manipulation. An illustration of these steps is provided in Fig.~\ref{fig: teaser}.

Addressing the first challenge involves contending the physical interactions between parts and the inherent ambiguities. The former is due to the collision between parts that prevent arbitrary assembly order, and the latter arises due to multiple viable assembly sequences. Can we learn the order of assembling the parts statistically from their geometry and their locations in the target assembly? For example, it is clear in Fig.~\ref{fig: teaser} that the red screw can only be inserted if the orange piece is in place -- the target blueprint and individual pieces collectively establish a specific order for assembly. We use this insight towards designing an implicit neural planning network using Transformers~\cite{transformer}, dubbed \emph{Part Assembly Sequence Transformer} (PAST) that takes as input point clouds of the target blueprint and the assembled parts and identifies the next parts to be assembled. Then, it is applied to generate the full sequence in an autoregressive fashion. For training our PAST model, we construct a benchmark dataset for part assembly sequences, dubbed D4PAS, by enumerating feasible assembly sequences~\cite{tian2022assemble}.

To solve the problem of part motion planning for assembly, we leverage the RRT-connect~\cite{rrt_connect} to generate trajectories from each part's resting pose to its assembled pose. Concurrently, we conduct an efficient physics-inspired multi-scale optimization of potential contact points on the part's surface to identify those that are most effective in achieving the desired part movement.
Upon generating the assembly plan through the aforementioned steps, prior research has explored the use of reinforcement learning~\cite{9812312, 8794127, dasari2023pgdm}, model-predictive control~\cite{10160649}, and diffusion policies~\cite{chi2023diffusionpolicy} for its execution. However, the focus of this work is not on physical execution, which is earmarked for future investigation.

In summary, our primary contributions are as follows:
\begin{itemize}
  \item We present an assembly planning algorithm to generate feasible part assemblies based on target blueprints, including inference of assembly sequences, planning of part movements, and optimization of contact points.
  \item We introduce the Part Assembly Sequence Transformer (PAST) to infer assembly sequence in an autoregressive fashion. PAST is designed to generalize to novel, diverse, and practical blueprints and part geometries.
  \item We provide a dataset for part assembly sequences (D4PAS), replete with assembly trajectories, enumerated assembly sequences, and viable contact points, thereby providing a foundation for future studies in robotic assembly. 
\end{itemize}


\section{Related Works}
\label{sec: related_works}

Previous research on \textbf{part assembly} varies in focus. On the one hand, approaches like~\cite{li2023general, li2020impartass} tackle the problem via blueprint segmentation and part pose regression, employing point classification and optimization. Huang et al~\cite{HuangZhan2020PartAssembly} use dynamic graph networks for geometric learning, while~\cite{NSM} utilize a shape-mating discriminator for shape-mating tasks.
On the other hand, reinforcement learning (RL) has shown promise in physical assembly executions~\cite{narang2022factory}. However, these RL methods often focus on fixed or single-category targets~\cite{9482981,9811973, Ghasemipour2022BlocksAL,yu2021roboassembly} and entail costly training~\cite{Ghasemipour2022BlocksAL, heo2023furniturebench}.

To simplify part assembly, some works that streamline \textbf{assembly sequence inference} focus on optimizing the assembly order of individual parts~\cite{rashid2012review}. Techniques leveraging precedence relationships in CADs have been used~\cite{su2009hierarchical}. The assembly-by-disassembly strategy has been highlighted for its efficiency~\cite{de1989correct, tian2022assemble}. In cases where parts are rigid, this strategy simplifies planning by reversing disassembly sequences~\cite{ghandi2015review}, leveraging the bijection between assemblies and disassemblies. However, these methods often rely on time-consuming physical simulations. To address this, our work introduces the Part Assembly Sequence Transformer (PAST) for efficient sequence reasoning and geometrical understanding of parts and blueprints. We train the network using a dataset, D4PAS, generated from GPU-based simulations~\cite{makoviychuk2021isaac}, employing assembly-by-disassembly techniques.

After sequence determination, robots require further guidance on \textbf{part motions and contact points} for efficient and robust execution. Works by~\cite{dasari2023pgdm, thomas2018learning} demonstrate the importance of object movements and contact locations for robotic manipulations. For part motion planning, sampling-based methods like Rapidly-exploring Random Trees (RRT~\cite{rrt_connect}) have proven effective in robotic motion planning. Zhu et al.~\cite{zhu2023difflfd} shows that contact optimization reliably identifies contact points for dexterous manipulation tasks. Building on these insights, our work fulfills part assembly with part motion generation and contact points identification, leaving the physical execution for future research.


\section{Problem Overview}
\label{sec: problem_overview}

This work employs a part assembly formulation consistent with~\cite{tian2022assemble, yu2021roboassembly}, as shown in Fig.~\ref{fig: teaser}. Given $M$ part meshes $\mathcal{M}=\{ \mathcal{M}_i \}_{i=1}^M$ and their respective 6D assembled poses in the target blueprints $p^{tgt}=\{ p_i^{tgt} \}_{i=1}^M$, the algorithm plans the assembly trajectories $( p^0, p^1, ...)$ for each part from their resting poses $p^0 = \{ p_i^0 \}_{i=1}^M$. We use $p_i^t$ to represent the pose of part $\mathcal{M}_i$ at time $t$ and use $p^t$ to represent poses of all parts at time $t$.
The algorithm assumes, at each time step, that only one part is in motion while the others remain stationary~\cite{tian2022assemble}.
The work breaks down the complex task of assembly planning into three sub-tasks: assembly sequence inference, part motion planning, and contact point selection.

For one possible assembly, let the assembly sequence be denoted by $m = ( m_k )_{k=1}^M$, specifying the order of part assembly. Each $m_k$ belongs to the set $\mathcal{M}$ and identifies the $k$th part to be assembled. Part movements are represented by $\mathcal{T} = ( p_{m_k} )_{k=1}^M$, where $p_{m_k}$ details the trajectory of parts when part $m_k$ is moving throughout its moving horizon. Contact points facilitating these movements $p_{m_k}$ are indicated by $\mathcal{C} = ( c_{m_k} )_{k=1}^M$. The assembly planning problem is formulated as:
\begin{equation*}
\begin{aligned}
 \mathbb{P}(p^0, p^1, ..., p^{tgt}) 
  = & \mathbb{P}(m, \mathcal{T}, \mathcal{C}) \\
  = & \mathbb{P}(m) \cdot \mathbb{P}(\mathcal{T} | m) \cdot \mathbb{P}(\mathcal{C} | m, \mathcal{T}),
\end{aligned}
\end{equation*}
\noindent where in this work, we assume a multi-level solution approach by sequentially solving for the sub-problems of (i) assembly inference, (ii) motion planning, and (iii) contact selection, in that order.

It is crucial to recognize that feasible assembly sequences are a subset of all possible part permutations. This constraint arises from the potential for part collisions, which precludes arbitrary assembly sequences. For instance, a washer must be in place before tightening a screw. To address this combinatorial inference problem approximately, we introduce the Part Assembly Sequence Transformer (PAST) to learn statistical correlations between the parts and the target blueprint to produce physically viable assembly sequences $\mathbb{P}(m)$. The network ingests both target blueprints and unassembled parts, outputting the next feasible parts for assembly. This iterative process determines the full assembly sequence, as in Fig.~\ref{fig: network}.

Upon establishing the assembly sequence $m$, part movement can be planned using established motion planning algorithms $\mathbb{P}(\mathcal{T} | m)$. Following that, the algorithm optimizes the contact points that the robot can utilize to execute these movements, $\mathbb{P}(\mathcal{C} | m, \mathcal{T})$, thus culminating in a coherent assembly process. 

\section{Assembly Planning}
\label{sec: seq_infer}

The preceding section outlines our multi-level approach to assembly planning. This section delves into the details of each of the three planning levels. 

\subsection{Part Assembly Sequence Transformer (PAST)}
\label{subsec: transformer}
\begin{figure}[tb]
\begin{center}
\includegraphics[width=3.0in]{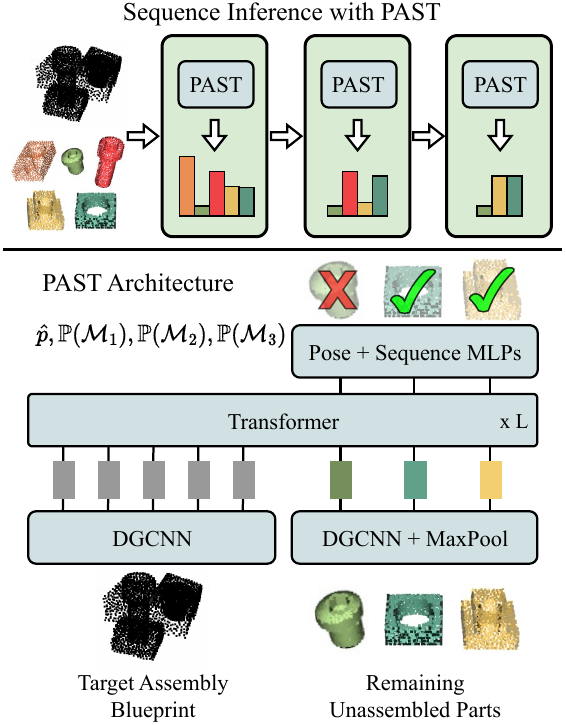}
\caption{
Sequence inference pipeline and PAST architecture.
(Top) PAST operates sequentially to estimate the assembly probability $\mathbb{P}(\mathcal{M}_i)$ for each remaining part. The part with the highest probability is chosen for assembly.
(Bottom) Using the third block as an example, PAST selects one part for assembly from the three remaining options. 
PAST also performs pose regression for each part $\hat{p}$ as an auxiliary task.
}
\label{fig: network}
\vspace{-0.5em}
\end{center}
\end{figure}

The transformer ingests the target assembly blueprint and the remaining unassembled parts, outputting a probability for each part's suitability for assembly at the current step. The part with the highest probability is chosen for assembly and removed from the list of remaining parts. This iterative process continues until all parts are assembled, yielding one assembly sequence $m = ( m_k )_{k=1}^M$. Fig.~\ref{fig: network} illustrates this recursive planning approach. 

PAST takes two branches of inputs: a target assembly blueprint and unassembled remaining parts, both represented using point clouds. For an assembly comprising $M$ parts, the target blueprint is rendered with the 6D assembled poses $p^{tgt}$ of all parts, resulting in a point cloud $PC_{tgt} \in \mathbb{R}^{N_t \times 6}$. This point cloud consists of $N_t$ sampled points, each with positional and normal features. During the $k$th step of assembly, PAST selects the next part to assemble from the $M-k$ remaining parts. These remaining parts are input as $M-k$ individual point clouds, denoted as $\{ PC_{r, i} \}_{i=1}^{M-k}$, where $PC_{r, i} \in \mathbb{R}^{N_r \times 6}$ and contains $N_r$ sampled points.

A key design question for PAST is which neural model to use for representing the input point clouds. Among point cloud encoders, such as PointNet and its variants~\cite{qi2017pointnet, qi2017pointnet++}, it was shown in~\cite{NSM} that dynamic graph CNN (DGCNN)~\cite{dgcnn} offers superior efficiency and representational capabilities in assembly segmentation.  To this end, we employ DGCNN to derive target features $v\in \mathbb{R}^{N_t \times h}$ from the target blueprint (one feature for every sample point) and part features $u_i \in \mathbb{R}^{h}$ from each remaining part (i.e., one feature for every part after max-pooling the features from all samples belonging to that part). 
Here, $h$ is the hidden feature dimension and $i \in \{1,..., M-k\}$.

Once the features are extracted, our PAST model then jointly refines these features through $L$ transformer blocks, as illustrated in Fig.~\ref{fig: network}.  As is well-known, transformers use self- and cross-attention to learn correlations between their inputs and have demonstrated state-of-the-art performances in generating sequential outputs (e.g., language). Our key insight is to use such attention to learn physically plausible assembly sequences in a supervised setting. Mathematically,  suppose for a \emph{query} set $\mathbf{q}$ and a \emph{key} set $\mathbf{k}$, let the transformer dot-product attention\footnote{For simplicity of notation, we have avoided details on multi-head attention and other processing modules in the transformer.} operator be defined as $\textrm{Attention}(\textbf{q}, \textbf{k}) = W_v(\textbf{k})^T \textrm{softmax} \left (   \frac{W_k(\textbf{k}) W_q(\textbf{q})}{\sqrt{h}} \right )$, where $W_q, W_k, W_v$ are matrices embedding the query and the key sets in a common latent space. 

Our PAST transformer blocks utilize a two-stage approach for feature refinement between the target and parts. The first stage independently processes and updates the features with self-attention; that is, 
\begin{equation*}
    v=\textrm{Attention}(v, v) \text{ and } u_i=\textrm{Attention}(U, u_i),
\end{equation*}
where $U=\{ u_i \}_{i=1}^{M-k}$ denotes all part point cloud features. 
The second stage applies cross-attention between the target features and part features~\cite{chen2021crossvit}, updating them to:
\begin{equation*}
    \hat{v}=\textrm{Attention}(U, v) \text{ and } \hat{u}_i=\textrm{Attention}(v, u_i).
\end{equation*}
where $\hat{v}$ and $\hat{u}_i$ are fed into the next block.\footnote{We use the same attention expression for all blocks except for replacing $u$ and $v$ updated to $\hat{u}$ and $\hat{v}$ from the previous block.}
In the final step, PAST calculates the assembly probability for each part using the formula
$\mathbb{P}(\mathcal{M}_i) = \textrm{MLP}(u_i)$.
The part to be assembled next is then selected based on the maximum probability, denoted as $m_k=\argmax_i  \mathbb{P}(\mathcal{M}_i)$.
In addition to predicting the assembly sequence, the transformer also estimates the 6D assembled pose $p^{tgt}_i$ for each part as an auxiliary task, represented as $\hat{p}_i = \textrm{MLP}_{p}(u_i)$. 
The inclusion of this auxiliary task enhances the network's capability to comprehend the geometric interrelations between parts, a technique that has proven effective in~\cite{NSM, li2023general}.

We use supervised learning for training PAST via estimating the predicted assembly probability $\mathbb{P}(\mathcal{M}_i)$ with mean squared error loss $\textrm{MSE} (y_i, \mathbb{P}(\mathcal{M}_i)) $ with the feasibility of assembling a part $\mathcal{M}_i$ at a given step is denoted $y_i$. Additionally, the pose regression task aims to minimize the difference between the actual 6D assembled pose $p^{tgt}_i$ and the predicted pose $\hat{p}_i$. The difference is calculated with $\sum_{i} \left \| p^{tgt}_{i,tra} - \hat{p}_{i,tra} \right \| + \left \| p^{tgt}_{i,rot} - \hat{p}_{i,rot} \right \|$, where $p^{tgt}_{i,tra}, \hat{p}_{i,tra}$ indicate target and predicted translation for each part and $p^{tgt}_{i,rot}, \hat{p}_{i,rot}$ represent the axis-angle.

\subsection{Dataset for Part Assembly Sequences (D4PAS)}
\label{subsec: dataset}

\begin{figure}[tb]
\begin{center}
\includegraphics[width=3.2in]{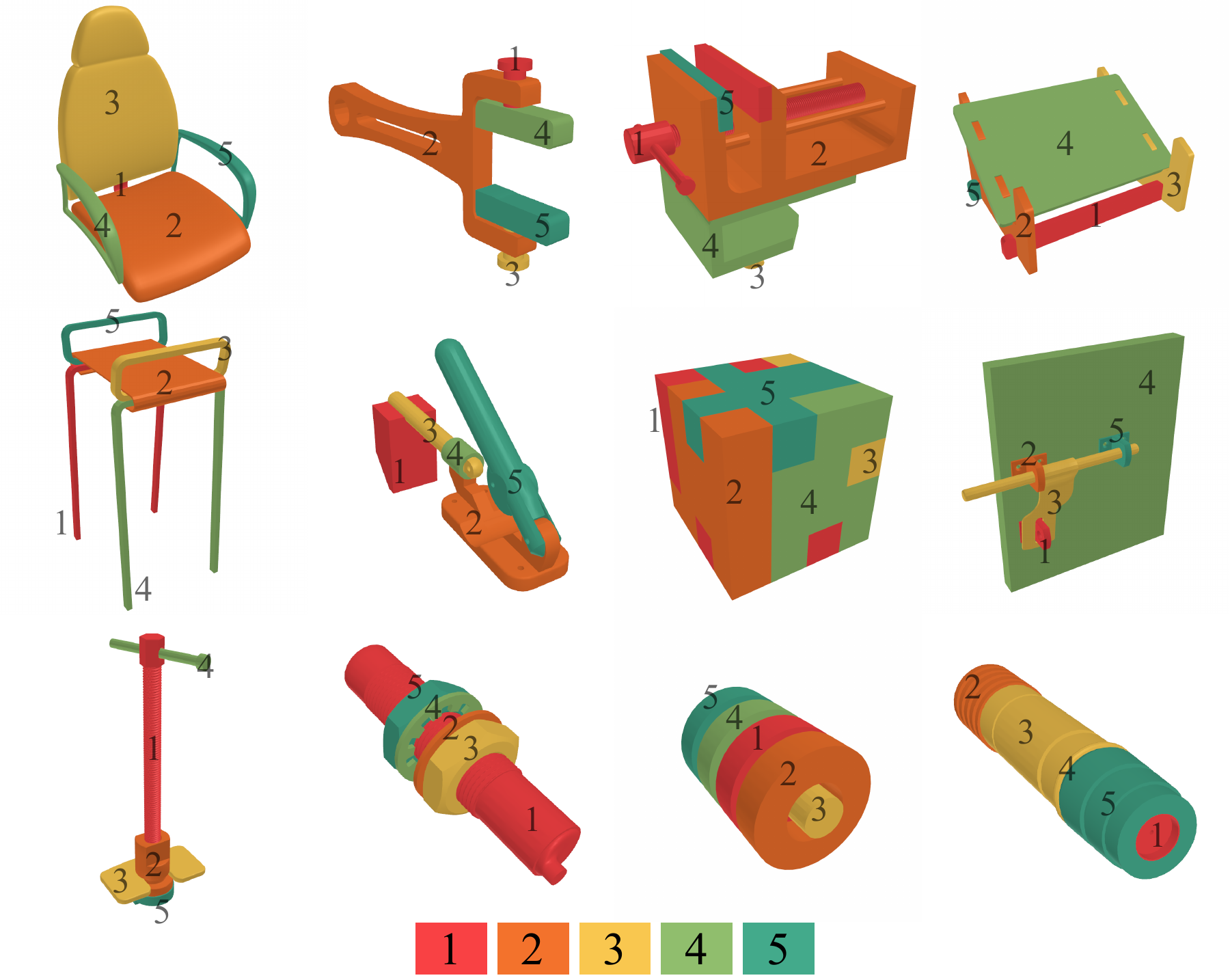}
\caption{Example assembly sequences in our dataset. For each target blueprint, we enumerate all feasible assembly sequences and present one representative sequence here. The color coding and the numbers beside each part signify the assembly sequence, as indicated by the color bar.
}
\label{fig: dataset}
\vspace{-0.5em}
\end{center}
\end{figure}

To train PAST, we introduce a new \emph{dataset for part assembly sequences} or D4PAS. Each sample in the assembly sequence dataset comprises multiple components, namely: (i) the target blueprint point cloud $PC_{tgt}$, (ii) point clouds for $M-k$ remaining parts $\{ PC_{r, i} \}_{i=1}^{M-k}$, and (iii) the feasibility of assembly for each part $\{ y_{i} \}_{i=1}^{M-k}$. The feasibility $y_{i}$ specifies whether a part $\mathcal{M}_i$ can be assembled at the current step and can subsequently lead to a successful final assembly. Note that there can be many viable part candidates at every step, derived from all possible sequence enumerations using the scheme in assembly-by-disassembly~\cite{tian2022assemble}, as described in Algorithm~\ref{algo: disassemble} and illustrated in Fig.~\ref{fig: dataset}.

\vspace{-0.5em}
\begin{algorithm}
   \caption{Disassembly Planning}
   \label{algo: disassemble}
    \begin{algorithmic}[1]
    
        \State \textbf{Input:} part meshes $\{ \mathcal{M}_i \}_{i=1}^M$ and inertias $\{ I_i\}_{i=1}^M$
        \State \textbf{Input:} target part pose $p^{tgt}$, empty queue $\mathcal{J}$
        \State \textbf{Output:} sequence of disassembly
        \State $\mathcal{J}.\texttt{enq}(p^{tgt}, f_{i}^j)$ for $f_{i}^j \propto I_i$ and $i\in \{1, \cdots,  M\}$ 

        \While{not finish} \Comment{In parallel}
            \State $(p^t, f_{i}^j)=\mathcal{J}.\texttt{deq}$ \Comment{BFS}
            \If{\texttt{success}($p^t$)}
                \State return $\texttt{GetSequence}(p^{tgt}, p^t)$
            \EndIf
            \State $p^{t-1}=\texttt{simulate}(p^t, f_{i}^j)$ \Comment{Disassembly attempt}
            \If {$\texttt{isNovel}(p^{t-1})$ and $\texttt{isExec}(p^{t}, p^{t-1})$}
                \State $\mathcal{J}.\texttt{enq}(p^{t-1}, f_{i}^j)$ for all unassembled part $i$
            \EndIf
        \EndWhile
\end{algorithmic}
\end{algorithm}
\vspace{-0.5em}

The disassembly planning algorithm operates in a search-based manner, where the set of parts poses' at time $t$, $p^t=\{ p_i^t \}_{i=1}^M$, serves as the search state. At each step, the algorithm selects an unassembled part $i$ (line 6) and attempts to remove it from the blueprint with a physical simulation (line 10). 

The selection of the moving part follows a breadth-first search (BFS) scheme, ensuring a comprehensive enumeration of all possible disassembly sequences.
The chosen part $i$ is moved in the direction of its moment of inertia $I_i$ with force $f_i^j$, calculated as $f_i^j = \Lambda_i e_j I_i$, where $\Lambda_i$ is the mass of the part and $e_j$ is a one-hot vector with nonzero element at $j\in\{1,2,3\}$. Torque is also applied to enable rotational movement and is computed similarly to $f_i^j$. 
 
The search queue is expanded only if the disassembly attempt results in novel part poses and is executable by a robot (line 11). 
Part poses are regarded as a novel ($\texttt{isNovel}$) if they lead to a new location for one of the parts. The feasibility of execution ($\texttt{isExec}$) is confirmed by determining whether there exists collision-free grasp or push points on the part's surface. 
After each disassembly attempt, the remaining unassembled parts are added back to the queue, considering all possible dragging forces (line 12).

Compared to~\cite{tian2022assemble}, the dataset generation algorithm in this work incorporates several advancements. First, we employ a parallel simulation (Isaac Gym~\cite{makoviychuk2021isaac}) to expedite the dataset generation and allow future studies on assembly planning with RL~\cite{Ghasemipour2022BlocksAL}. Second, unlike~\cite{tian2022assemble} that use arbitrary force directions, our algorithm applies disassembly forces along the part's moment of inertia, aligning with the physical properties of the parts. Third, we incorporate additional constraints to ensure that the generated plans are executable by robots ($\texttt{isExec}$), making our dataset more practical for further robotic applications.

\subsection{Part Motion and Contact Planning}
\label{subsec: obj_move_contact_planning}

Once the sequence is determined by the PAST transformer, the next steps involve planning the movements of the parts and identifying suitable contact points for robotic manipulation.
For part movements, we use the RRT-connect~\cite{rrt_connect} to search for a collision-free path for each part, in line with the inferred assembly sequence.

For contact points planning, this work first plans robust grasps for each part with two contact points and filters out those in collision with other parts~\cite{9811685, 9561954}. To do this, we enumerate all grasp pairs from the part point cloud $PC_{r,i}$ and use the Ferrari Canny metrics to determine their robustness~\cite{ppojpo}. We then identify the feasibility of execution using FCL~\cite{fcl} to check the collision between grasp points and the assembled parts along the assembly trajectory. If no feasible grasps are found, we resort to optimizing for a single pushing point $c$ by solving the optimization~\cite{zhu2023difflfd}:

\begin{equation}
\label{eq: force_opt}
\begin{aligned} 
\min_{c, F_c} & \quad F_{c} \\
s.t.  & \quad F_c \in FC(c) \\
 & \quad G(c) F_{c} = W
\end{aligned}
\end{equation}

\noindent where $F_c$ is the pushing force at $c$. $FC(c)$ is the friction cone at point $c$ and is expressed by $F_{c, 1}^{2} + F_{c, 2}^{2} \leq \mu F_{c, 3}^{2}$ with $\mu$ being the friction coefficient. $G(c)\in \mathbb{R}^{6\times 3}$ is the grasp map which maps the contact force to the part movement force~\cite{force_opt}. $W$ is the part movement force derived from the part movement~\cite{zhu2023difflfd}.
We observe that the optimization problem~\eqref{eq: force_opt} can be cast as a semi-definite program (SDP) once the pushing point $c$ is specified. To solve this problem, we employ a hybrid approach that combines sampling with an SDP solver~\cite{zhu2023difflfd}: the pushing point $c$ is sampled from the part's point cloud and held constant during the optimization of $F_c$ as SDP. The outcomes are iteratively updated across all sampled pushing points to identify the optimal solution.


\section{Experiments}
\label{sec: exp}

This section offers details of our model, dataset, and empirically validates our approach against the related methods. 

\subsection{Dataset and Network Details}
\label{subsec: details}

Both disassembly planning and PAST training were performed using a single RTX3090 GPU. During the disassembly planning, properties of parts $\Lambda_i, I_i$ were extracted from the simulation. The friction $\mu=0.2$. 

Our dataset comprises 8,670 target blueprints and a total of 84,326 assembly sequences, broken down as follows: 7,278 are 3-step sequences, 54,612 range from 4 to 7 steps, and 22,436 have more than seven steps.
These sequences can be further augmented with varying choices of the remaining (unassembled) parts. Specifically, for an assembly sequence with $M$ parts, we randomly split the sequence into two segments with size $k$ and $M-k$, respectively. Parts in the second segment are regarded as unassembled, and the first part in the second segment is the part that can be assembled with $y_{M-k}=1$. Additionally, we gather segments from all viable sequences to identify all potential parts that can be assembled for a specific unassembled segment.

During the training of PAST, two-part assemblies were only used to train the auxiliary pose regression. The target blueprint was sampled at $N_t = 1024$ points, and each part mesh was sampled at $N_r = 512$ points. The DGCNN encodes point clouds with dimension $h=256$. PAST uses $L=8$ transformer blocks and was implemented in PyTorch and was trained with the AdamW optimizer using the One-Cycle learning rate scheduler. The target blueprints were re-centered and normalized to a unit ball and randomly rotated and jittered as augmentation. We allocate 500 multi-part assemblies for the test set, reserving others for training.

\subsection{Baselines and Ablations}
\label{subsec: baseline_ablation}

The baseline and ablation studies aim to evaluate the efficacy of PAST and the overall algorithm.

\noindent\textbf{Metrics.} We employ two key metrics to assess the performance of assembly sequence inference: one-step prediction accuracy (\texttt{1-Acc}) and sequence prediction accuracy (\texttt{Seq-Acc}).
A one-step prediction is deemed correct if the selected part, sampled based on predicted assembly probabilities, belongs to the set of possible parts for assembly.
In addition, we apply PAST in an autoregressive fashion to generate a full assembly sequence, which is considered correct if it aligns with any of the possible assembly sequences for that object in our dataset.
In addition, we also report the computational time (CT) taken for full sequence inference.

\noindent\textbf{Compared algorithms.} Given that no existing solutions are tailored specifically for part assembly sequence inference, we adapt methods from part segmentation and pose regression as baselines, and compare against various ablations.

\begin{itemize}
  \item NSM (Neural Shape Mating~\cite{NSM}) uses a transformer to address two-part shape mating. We adapt this network to accommodate multiple part inputs.
  \item DGL (Dynamic Graph Learning~\cite{DGL}) employs a graph neural network to perform assembly pose regression. We use a global node to represent target assembly~\cite{graphnet}, which facilitates assembly sequence inference.
  \item ATA (Assemble-Them-All~\cite{tian2022assemble}) solves assembly through runtime physical simulation. The assembly process aligns with Algorithm~\ref{algo: disassemble}.
  \item Seg-PAST: We substitute the final pose regression layer in PAST to predict blueprint segmentation during the pretraining, as advised in \cite{li2023general}.
  \item NoAux-PAST: We eliminate the auxiliary pose regression task, focusing solely on predicting the assembly sequence using the same network.
\end{itemize}

\subsection{Experimental Results}
\label{subsec: exp_results}

\begin{table}[t]
\centering
\caption{Quantitative results on assembly sequence inference. We report three metrics: one-step prediction accuracy (\texttt{1-Acc}), full sequence prediction accuracy ($\texttt{Seq-Acc}$), and the computation time (CT).}
\label{tab: exp}
\vspace{-0.5em}
\begin{tabular}{c|cccc}
\hline
            & \texttt{1-Acc} (\%)         & \texttt{Seq-Acc} (\%)        &  CT ($ms$)    \\ \hline
NSM~\cite{NSM}        & 75.0                        & 57.8                         & 58.5          \\
DGL~\cite{DGL}        & 77.4                        & 54.1                         & 104.4         \\
ATA~\cite{tian2022assemble}        & NA               & NA                & 24312.6       \\
Seg-PAST   & 90.3                        & 80.4                         & \textbf{52.2} \\
NoAux-PAST & 79.1                        & 58.4                         & \textbf{52.2} \\ \hline
PAST       & \textbf{91.7}               & \textbf{82.9}                & \textbf{52.2} \\ \hline
\end{tabular}
\vspace{-0.5em}
\end{table}

\begin{figure}[tb]
\begin{center}
\includegraphics[width=3.2in]{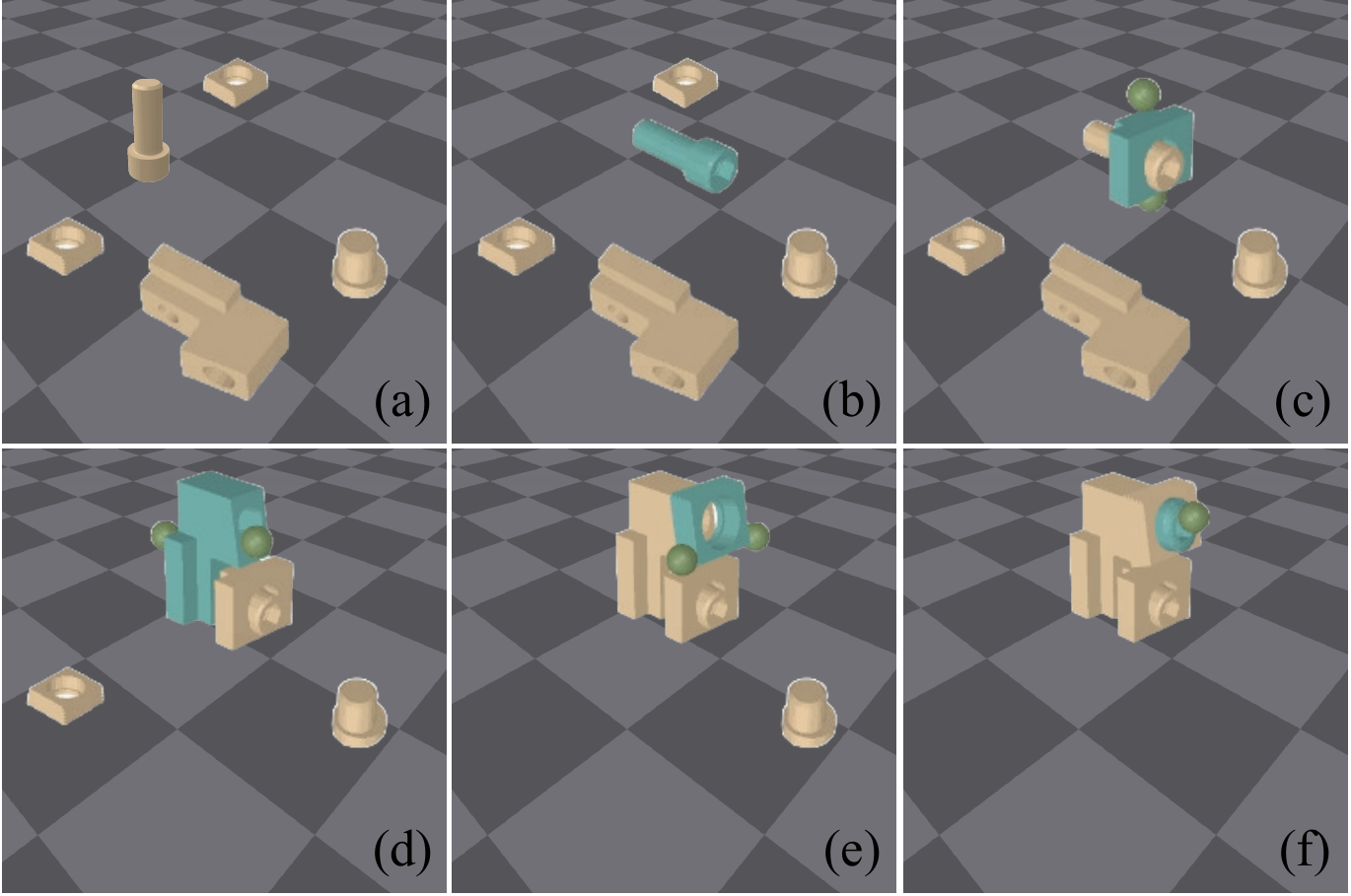}
\caption{Assembly planning results (a-f). In each step, the part colored blue indicates the one in motion, while the yellow parts signify those that are stationary. In (c, d, e), the dual green spheres denote feasible grasp points. In (f), a solitary green sphere highlights the designated pushing point.}
\label{fig: demo}
\vspace{-0.5em}
\end{center}
\end{figure}

\begin{figure}[tb]
\begin{center}
\includegraphics[width=2.6in]{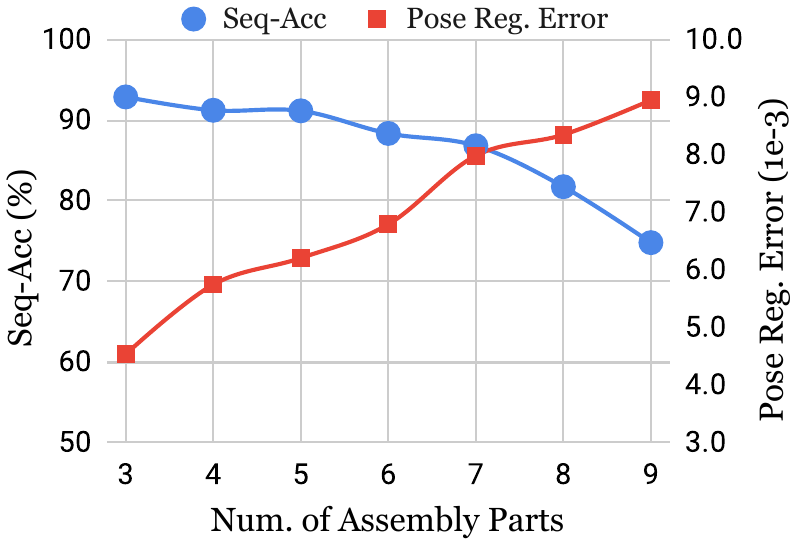}
\caption{Sensitivity analysis for sequence inference accuracy and auxiliary pose regression error.}
\label{fig: exps}
\vspace{-0.5em}
\end{center}
\end{figure}

Table~\ref{tab: exp}, Fig.~\ref{fig: demo}, and Fig.~\ref{fig: exps} summarize the quantitative and qualitative results, which are elaborated upon in this section. For more results, we refer readers to our video. 

\noindent\textbf{Multi-level assembly planning.} As outlined in Sections~\ref{sec: introduction} and~\ref{sec: problem_overview}, our methodology breaks down assembly planning into three distinct phases.
This approach achieved assembly planning in $1565.9$ ms, with an $82.9\%$ success rate across our 500 multi-part assemblies test set. Figure~\ref{fig: demo} illustrates an example of assembling parts from scratch. Specifically, part movement planning averaged $1476.9$ ms with a $100\%$ success rate, while contact point optimization took $36.8$ ms with $100\%$ success rate using multi-threaded computation and the CVXPY solver~\cite{diamond2016cvxpy}. 
Unlike our method, RL-based assembly planning~\cite{yu2021roboassembly} exhibits inferior performance when evaluated in our setting, achieving a 63.9\% assembly success rate. While prior works showed promise with simpler geometries~\cite{yu2021roboassembly, Ghasemipour2022BlocksAL}, we posit that end-to-end policies may struggle to handle complex geometries and assembly reasoning simultaneously.

\noindent\textbf{Generalization to novel assemblies.} From Table~\ref{tab: exp}, we see that PAST consistently outperforms other neural sequence inference models, such as NSM, DGL, Seg-PAST, and NoAux-PAST in both one-step and full-sequence prediction tasks. Unlike NSM, which employs a discriminator for target shape understanding and fails to capture the assembly geometries' distribution, PAST leverages the target assembly blueprint as input and can extract direct features from the target assembly. DGL, which represents parts as a graph and updates features at the node level, struggles to model geometries from other parts and the target shape. In contrast, PAST aggregates features at the point level, thus enhancing geometric understanding, consequently yielding superior learning outcomes.

Further, from Table~\ref{tab: exp}, we also see from the ablated PAST variants that incorporating auxiliary tasks, such as pose regression or part segmentation, significantly improves network performance. This improvement is attributed to the additional guidance these tasks offer, enhancing the network's understanding of part interactions, which are key for assembly sequence inference. Interestingly, target segmentation underperforms compared to pose regression, possibly because some points in the target assembly, like those corresponding to assembled parts, are not supervised during training. Further, as is expected, ATA takes significant computing as it uses enumeration and simulation for assembly sequences. Instead, PAST, which shows promising accuracy, has a dramatically short computing time, making it well-suited for real-world robotic assembly. In Figure \ref{fig: exps}, we analyze the impact of the number of parts on sequence inference accuracy and auxiliary pose regression error. The results indicate that increased part count leads to reduced sequence accuracy and higher pose error, corroborating results from~\cite{Ghasemipour2022BlocksAL}: the complexity of the assembly problem increases with the number of parts in the target blueprint.

\noindent\textbf{Disassembly Planning.} As noted earlier, while our D4PAS construction bears similarities to~\cite{tian2022assemble}, we enhance disassembly planning through: (i) employing GPU-accelerated simulation and (ii) applying part-centric forces. Enabling parallel computations thereby reducing compute time for disassembly planning to $3592.0$ ms against $24312.6$ ms. This efficiency not only allows for the enumeration of feasible assembly sequences within a manageable timeframe but also establishes a performance benchmark for future research in end-to-end robotic assembly learning~\cite{Ghasemipour2022BlocksAL}. Second, we apply part-centric disassembly forces aligned with the parts' inertia axes, as supported by findings in~\cite{willis2020fusion, Willis_2022_CVPR}. Although this adjustment in the force application coordinate system may seem minor, it led to a notable increase in planning success rate: $92.7\%$ in our approach versus $83.8\%$ in~\cite{tian2022assemble}.


\section{Conclusion}
\label{sec: conclusion}

This paper makes several key contributions to robotic assembly. First, we introduce a multi-level framework for generating assembly plans, encompassing part sequences, motions, and contact points. Second, we unveil the Part Assembly Sequence Transformer (PAST) for inferring feasible assembly sequences based on target blueprints and part geometries. Third, we offer a large-scale benchmark dataset for part assembly sequence (D4PAS) featuring thousands of physically validated sequences. Post-sequence inference, we employ motion planning and contact optimization to complete part assembly. Our evaluations show that PAST and the overall algorithm match previous simulation-based methods but with significantly reduced computation time. 

The present work has some limitations. Our approach assumes one moving part at a time, which may limit practical robotic execution. Additional mechanisms may be needed to hold the assembly. Currently, we don't address the robotic execution of the plans and assume ideal part sensing and localization. Future work will focus on fixing these limitations while enabling real-world robotic assembly and tool use.


\addtolength{\textheight}{-1cm}   

\bibliographystyle{IEEEtran}
\typeout{}
\bibliography{references}

\end{document}